\documentclass{llncs}
\usepackage{amsmath,amssymb}
\usepackage{graphicx}
\usepackage{booktabs}
\usepackage{microtype}
\usepackage{xcolor}
\usepackage[hidelinks]{hyperref}

\title{Reliable Egocentric Action Anticipation via Temporal Reliability Suppression and Compositional Graph Decoding}

\author{Mahsa Mohammadi \and Sareh Rowlands}
\institute{University of Exeter, Exeter, United Kingdom\\
\email{mm1510@exeter.ac.uk}, \email{S.Rowlands@exeter.ac.uk}}

\begin{document}
\maketitle

\noindent\textbf{Abstract.}
Wearable action anticipation systems~\cite{ego4d,affttention,egosurvey} must remain reliable despite missing frames, masking, and sensor noise~\cite{temprobust,stcorrupt}, yet existing egocentric anticipation methods largely assume clean observations.
We identify two complementary failure modes under temporal corruption: unreliable temporal evidence during encoding and implausible, low-support verb--noun compositions during decoding.
We address them with a lightweight framework combining Temporal Reliability Suppression~(TRS) and Robust Verb--Noun Graph~(RVG) decoding.
TRS predicts a per-frame suppression score from the projected input embedding and uses it as a learned key-side attention penalty at every encoder block and to derive reliability-weighted temporal pooling.
RVG re-ranks verb--noun pairs using a PMI-based compatibility graph constructed from training labels.
Under corruption-augmented training, TRS+CA+RVG reaches 29.1\% average corrupted accuracy and 88.2\% relative robustness across six corruptions, including three mechanisms absent during training, while reducing rare verb--noun predictions from 15.4\% to 1.1\%.
Multi-seed and diagnostic experiments show that TRS responds to synthetic masking; shuffled-graph and frequency-only controls further indicate that RVG gains depend on genuine pairwise compatibility rather than marginal-frequency effects alone.


\section{Introduction}

Wearable assistants must not merely recognise what a user is doing; they must anticipate what the user is \emph{about to do}~\cite{ego4d,egosurvey,affttention}.
This capability underpins daily-activity support, cooking assistance, and human--robot interaction, where even a short lead time can enable proactive assistance~\cite{rulstm,avt,affttention,egoagent}.

Despite rapid progress in egocentric anticipation, current methods implicitly assume that the observed video is reliable.
This assumption rarely holds for wearable devices, where motion blur, dropped frames, occlusion, bandwidth limitations, and sensor failures frequently degrade the visual stream.
For a proactive wearable assistant, such failures can trigger an intervention at the wrong time or recommend an inappropriate action, making robust anticipation a prerequisite rather than an optional deployment feature.

The field has advanced through temporal forecasting~\cite{rulstm,avt,hro,futr}, uncertainty-aware or semantic verb--noun modelling~\cite{uadt,sgear,insight,querymamba,parvla}, and structured egocentric representations~\cite{easg}, while practical streaming and resource-constrained perception introduce further deployment constraints~\cite{streaming,egoadapt}.
A parallel line of work shows that video models are sensitive to frame-level degradation~\cite{temprobust,stcorrupt,robustformer}, but primarily focuses on detection or generic recognition.
Our work lies at the intersection of egocentric action anticipation, robust video understanding, and compositional reasoning.
To our knowledge, we provide the first controlled study that jointly examines temporal-corruption robustness and verb--noun compositional failure in egocentric anticipation.

\noindent\textbf{Our observation.}
Corruption creates two coupled failures.
First, degraded frames provide \emph{unreliable temporal evidence}: an encoder with no explicit reliability signal may assign substantial attention to corrupted frames, producing unstable features.
Second, even when the encoder partially recovers, \emph{verb--noun composition can break}: individually plausible verb and noun scores may combine into a pair the training distribution almost never contains, such as \textit{open knife}.
We define \emph{low-support} pairs as verb--noun combinations with little or no training evidence, irrespective of whether they are physically possible.
The two failures are complementary: fixing temporal encoding does not guarantee plausible compositions, and enforcing compositional plausibility does not help when the temporal evidence is already corrupted.

\begin{figure}[t]
\centering
\includegraphics[width=0.98\linewidth]{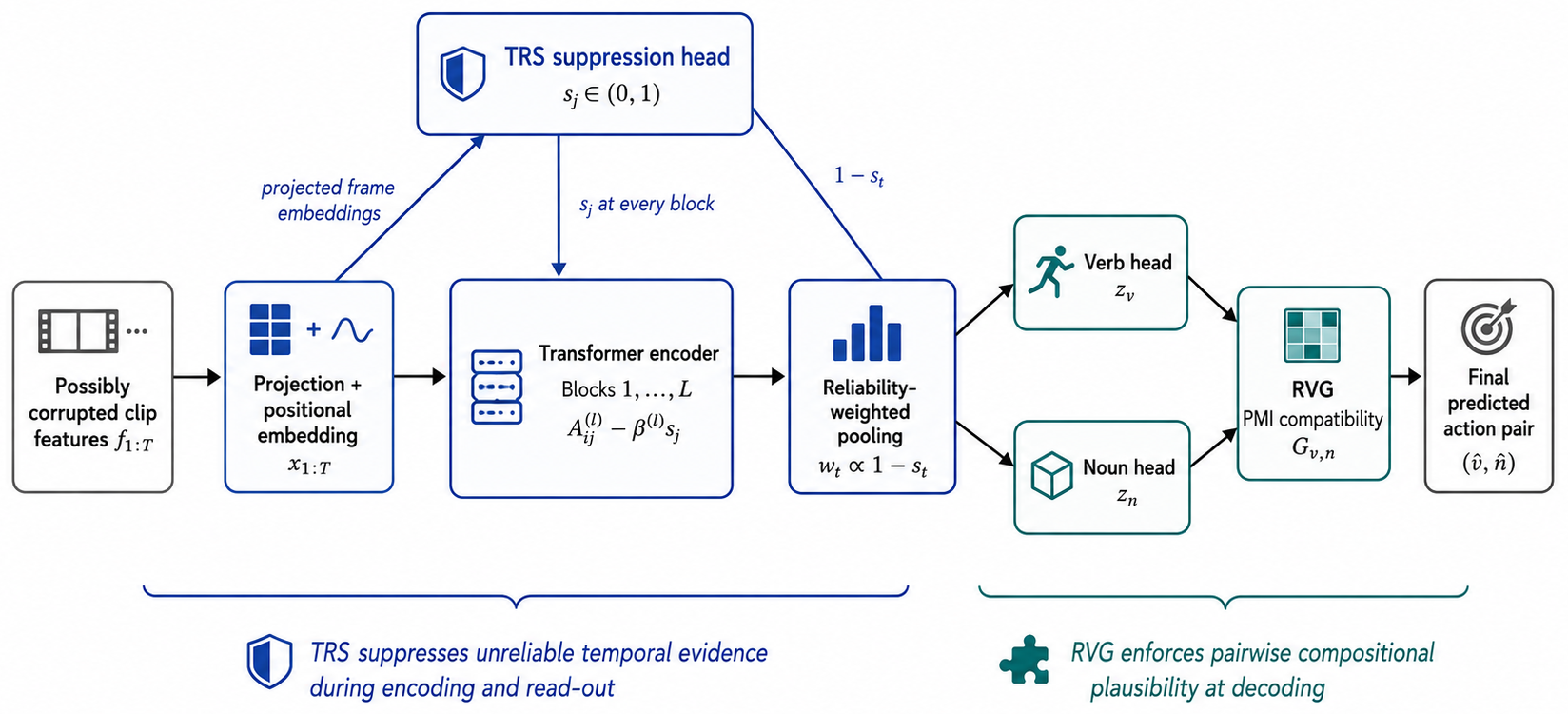}
\caption{Overview and complementarity of TRS+RVG. Temporal corruption affects egocentric anticipation at two levels: TRS suppresses unreliable temporal evidence during encoding and pooling, whereas RVG corrects low-support verb--noun compositions at decision time.}
\label{fig:pipeline}
\end{figure}

\noindent\textbf{Our approach.}
TRS+RVG addresses each failure with a dedicated lightweight module.
Temporal Reliability Suppression~(TRS) predicts a per-frame suppression score and uses it twice: as a key-side penalty inside every encoder block~\cite{vaswani2017attention}, and through $1-s_j$ as a reliability weight at temporal pooling.
Robust Verb--Noun Graph~(RVG) decoding re-scores pairs with a PMI-based~\cite{church1990word} compatibility graph from training labels.
Figure~\ref{fig:pipeline} summarises the full pipeline and the complementarity between representation-level suppression and decision-level correction.

\noindent\textbf{Contributions.}
\begin{itemize}
\setlength{\itemsep}{1pt}
\setlength{\parsep}{0pt}
\setlength{\topsep}{0pt}
\setlength{\partopsep}{0pt}
\item We identify two complementary failure modes in egocentric action anticipation under temporal corruption: unreliable temporal evidence during encoding and implausible low-support verb--noun compositions during decoding.
\item We propose a lightweight robustness framework, \textbf{TRS+RVG}, that preserves the backbone depth and feature dimensions without adding encoder blocks: TRS adds only $0.73\%$ parameters, while RVG performs lightweight post-hoc compositional correction.
\item We introduce a robustness evaluation protocol comprising six temporal corruption types, including three corruption mechanisms absent during training, together with a relative robustness metric and transfer evaluation on EPIC-Kitchens-100.
\item We provide multi-seed experiments, diagnostic studies, control ablations, and transfer experiments supporting the complementary roles of TRS and RVG: TRS responds to frame-level synthetic masking, whereas RVG gains depend on genuine pairwise compatibility rather than marginal label frequency.
\end{itemize}

\section{Related Work}

\paragraph{Egocentric action anticipation.}
RULSTM~\cite{rulstm} establishes multi-scale recurrent anticipation; AVT~\cite{avt}, HRO~\cite{hro}, and FUTR~\cite{futr} develop Transformer- and representation-forecasting alternatives.
Streaming evaluation~\cite{streaming}, efficient multisensory perception~\cite{egoadapt}, and unified predictive agents~\cite{egoagent} broaden the deployment setting.
UADT~\cite{uadt} decouples verb and noun prediction under uncertainty, while S-GEAR~\cite{sgear} learns semantically structured action prototypes.
None explicitly evaluates compositional anticipation under controlled temporal corruption; we target degraded wearable inputs.

\paragraph{Semantic verb--noun composition.}
INSIGHT~\cite{insight} uses verb--noun co-occurrence for long-term anticipation, QueryMamba~\cite{querymamba} introduces a statistical interaction module, and PAR-VLA~\cite{parvla} reasons over verb and noun prototypes.
Egocentric Action Scene Graphs~\cite{easg} further demonstrate the value of structured relational representations for downstream anticipation, while AntGPT~\cite{antgpt} supplies language-model priors.
Unlike these learned or long-horizon approaches, RVG is used as a corruption-robust decoding prior.

\paragraph{Robust video understanding.}
Li et al.~\cite{temprobust} benchmark temporal action detection under five frame-level corruptions, finding large drops even at mild severity and proposing FrameDrop augmentation; Yi et al.~\cite{stcorrupt} show spatial and temporal robustness are distinct; RobustFormer~\cite{robustformer} uses noise-robust masked-autoencoder pretraining, building on image common-corruption benchmarks~\cite{imagenetc}.
All target detection or recognition; we show that robustness for verb--noun \emph{anticipation} additionally requires plausible compositional decoding. Unlike RobustFormer and corruption benchmarks for recognition or detection, our setting evaluates anticipation, where corruption can affect both temporal evidence and the composition of independently predicted verb and noun labels. Accordingly, our contribution is not a new generic robustness backbone, but a task-specific analysis and lightweight correction framework for compositional egocentric anticipation.

\paragraph{Wearable and assistive egocentric AI.}
AFF-ttention~\cite{affttention} anticipates short-term object interactions for wearable assistance, and EgoAction~\cite{egoaction} studies reliability-aware fusion for egocentric \emph{detection}, noting verb and noun streams fail differently.
Plizzari et al.~\cite{egosurvey} identify the gap between laboratory benchmarks and practical wearable systems as a central open problem, motivating this study.

\section{Method}

\subsection{Problem Setup}

Given a clip $\mathbf{X}=\{x_t\}_{t=1}^{T}$ observed $\tau_a$ seconds before action onset, the goal is to predict the next action as a verb--noun pair $(v^*,n^*)\in\mathcal{V}\times\mathcal{N}$.
Both datasets use a 10-second observation window with $T=10$ pre-extracted TSN feature tokens; $\tau_a=0.5$ seconds for EGTEA and $1.0$ second for the EPIC transfer split.
All corruptions are applied at the feature level rather than to raw pixels.

\paragraph{Corruption protocol.}
During training, selected tokens receive additive feature noise, zero masking, or local temporal smoothing, and the selected positions define the frame-level corruption labels used to supervise TRS.
Approximately 60\% of training clips are corrupted, while the remaining clips are kept clean.
The main evaluation uses six high-severity corruptions: Mask, Noise, Combined, Drop, Jitter, and Freeze.
Combined sequentially composes the three training operators but is not sampled as a distinct training condition; Drop, Jitter, and Freeze use mechanisms absent during training.
We additionally evaluate standalone Blur as severity transfer because temporal smoothing is seen during training at a smaller kernel width.
A fixed evaluation seed gives every method the same corruption realisation.
The exact training and evaluation operators, boundary handling, control implementations, and reproducibility settings are specified in Sec.~\ref{sec:repro_details}.

\subsection{Temporal Reliability Suppression (TRS)}

\paragraph{Motivation.}
A corrupted frame does not merely contribute bad evidence of its own; through self-attention it can contaminate neighbouring representations before the encoder has an opportunity to recover.
TRS therefore estimates frame reliability before any temporal mixing and uses the estimate wherever frame information is aggregated.

\paragraph{Suppression estimation.}
We build on an $L$-layer pre-norm Transformer encoder~\cite{vaswani2017attention} with $H$ heads and width $d$.
Within each head, scaled dot-product attention at block $\ell$ is
\begin{equation}
A^{(\ell)}_{ij}=\frac{Q^{(\ell)}_i{K^{(\ell)}_j}^{\top}}{\sqrt{d_k}},
\qquad d_k=d/H.
\label{eq:attn}
\end{equation}
TRS predicts a per-frame suppression score from the projected input embedding $x_j=W_{\mathrm{in}}f_j+p_j$:
\begin{equation}
s_j=\sigma\!\left(W_2\,\mathrm{GELU}(W_1x_j+b_1)+b_2\right),
\qquad s_j\in(0,1).
\label{eq:head}
\end{equation}
The head is evaluated once before the first encoder block; $s_j\approx1$ denotes an observation estimated to be corrupted, while $1-s_j$ denotes estimated reliability.
The head reads each frame independently and therefore has no explicit temporal context. In the implementation, $W_1$ maps $512\rightarrow128$, followed by GELU and Dropout$(0.1)$, and $W_2$ maps $128\rightarrow1$.

\paragraph{Two complementary uses.}
During attention, the suppression score reduces the influence of unreliable keys:
\begin{equation}
\tilde{A}^{(\ell)}_{ij}=A^{(\ell)}_{ij}-\beta^{(\ell)}s_j,
\qquad
P^{(\ell)}_{ij}=\mathrm{softmax}_j\!\left(\tilde{A}^{(\ell)}_{ij}\right).
\label{eq:bias}
\end{equation}
Because the penalty depends on key index $j$, an unreliable observation receives less attention from every query.
Each block uses $\beta^{(\ell)}=\mathrm{softplus}(\tilde\beta^{(\ell)})$ to enforce non-negativity, with $\tilde\beta^{(\ell)}=\log(e^2-1)$ so that $\beta^{(\ell)}=2.0$ initially. Padded keys are still assigned $-\infty$ by the standard attention mask; the finite TRS penalty does not replace padding exclusion.

At read-out, the same scores derive reliability-weighted pooling:
\begin{equation}
w_t=\frac{1-s_t}{\sum_{t'=1}^{T}(1-s_{t'})+\varepsilon},
\qquad
\bar{h}=\sum_{t=1}^{T}w_t h_t^{(L)},
\qquad \varepsilon=10^{-6}.
\label{eq:pool}
\end{equation}
The verb and noun classifiers operate on $\bar{h}$.
Thus, TRS limits corruption while representations are formed and again when they are aggregated for prediction. Padded positions are assigned $s_t=1$ during pooling. The all-suppressed case was not observed in the reported runs; $\varepsilon$ is retained for numerical stability.

\paragraph{Supervision.}
The suppression head is trained with the binary corruption labels $y_j$ produced by the training corruptor:
\begin{equation}
\mathcal{L}_{\mathrm{TRS}}
=-\frac{1}{|\Omega|}\sum_{j\in\Omega}
\left[y_j\log s_j+(1-y_j)\log(1-s_j)\right],
\label{eq:loss_trs}
\end{equation}
where $\Omega$ contains non-padded positions.
The full objective is
\begin{equation}
\mathcal{L}=\mathcal{L}_v+\mathcal{L}_n+\beta_{\mathrm{sup}}\mathcal{L}_{\mathrm{TRS}},
\qquad \beta_{\mathrm{sup}}=0.5.
\label{eq:loss_total}
\end{equation}
At inference no corruption labels are provided: $s_j$ is predicted directly from the input features.

\subsection{Robust Verb--Noun Graph Decoding (RVG)}

Even when temporal encoding succeeds, independently predicted verbs and nouns may combine into low-support actions under degraded evidence.
Rather than retraining the classifier, RVG corrects these compositions during decoding using a PMI-based compatibility graph constructed from training labels.
This complements TRS: TRS improves the reliability of temporal evidence, whereas RVG corrects the semantic structure of the final prediction.

The verb and noun heads produce independent log-probabilities $z_v\in\mathbb{R}^{|\mathcal{V}|}$ and $z_n\in\mathbb{R}^{|\mathcal{N}|}$.
Using raw empirical probabilities, we define the shrunk PMI graph
\begin{equation}
G_{v,n}=\frac{N_{v,n}}{N_{v,n}+k}\log\frac{P(v,n)}{P(v)P(n)},
\qquad k=10,
\label{eq:spmi}
\end{equation}
clipped to $[-3,3]$.
Pairs with $N_{v,n}<5$ receive $G_{v,n}=0$ without evaluating the logarithm.
The final prediction is
\begin{equation}
S(v,n)=z_v(v)+z_n(n)+\lambda G_{v,n},
\qquad
(\hat v,\hat n)=\arg\max_{v,n}S(v,n).
\label{eq:score}
\end{equation}
The graph uses training labels only, and $\lambda=2$ is selected on the EGTEA validation split and transferred unchanged to EK100.
At $\lambda=1$, ignoring shrinkage and clipping, Eq.~\eqref{eq:score} replaces the independent prior implicit in the decoupled heads with the train-set joint prior.

\section{Experiments}

\subsection{Setup}

\paragraph{Datasets.}
\textbf{EGTEA Gaze+}~\cite{egtea}: 19 verbs, 51 nouns, 106 unique actions; our processed split-1 partition contains 10,321 clips.
\textbf{EPIC-Kitchens-100}~\cite{epickitchens100}, extending EPIC-Kitchens~\cite{epickitchens}: our feature pipeline yields 125 verb and 352 noun indices.\footnote{These cardinalities come from the pre-extracted RULSTM-style feature/annotation pipeline we reuse, which retains a larger index space than the official EK100 taxonomy of 97 verbs and 300 nouns. Because of this and our video-disjoint 90/10 split of the public training videos, our EK100 numbers are not comparable to the official challenge leaderboard; we use EK100 only for within-pipeline comparison.}

\paragraph{Features and training.}
We use pre-extracted 1024-d TSN RGB features~\cite{tsn} following the RULSTM protocol~\cite{rulstm}.
The backbone is a 4-layer pre-norm Transformer with $d=512$, $H=8$ heads, feed-forward width 1024, dropout 0.1, and learnable positional embeddings.
We train with AdamW (learning rate $10^{-4}$, weight decay $10^{-4}$), gradient clipping at 1.0, and early stopping with patience 5 on validation joint-action accuracy.
We report top-1 action accuracy, requiring both verb and noun to be correct.

\paragraph{Calibration, metrics, and cost.}
The compatibility graph is constructed exclusively from training labels. The graph weight $\lambda$ is selected on the EGTEA validation split from $\{0,0.5,1,2\}$ using mean top-1 joint-action accuracy over Mask, Noise, and Combined; test labels are never used for graph construction or hyperparameter selection. Standalone Blur is excluded from calibration. The resulting $\lambda=2$ is fixed for all reported comparisons and transferred to EK100 without re-tuning.
AvgC averages the six main corruptions, and Relative Robustness is $\mathrm{RR}=100\times\mathrm{AvgC}/\mathrm{Acc}_{\mathrm{clean}}$~\cite{temprobust}; RR is interpreted alongside absolute accuracy.
TRS adds 65,797 parameters, or $0.73\%$ of the 9.04M-parameter model.
RVG adds no trainable parameters; on EGTEA it stores a $19\times51$ compatibility table and evaluates 969 pair scores.

\subsection{Corruption, Control, and Reproducibility Details}
\label{sec:repro_details}

\paragraph{Training corruption.}
A batch is routed through the corruptor with probability $0.7$, and each routed
clip remains clean with probability $0.15$, giving an effective corrupted-clip
rate of $0.7\times0.85=0.595$. For a corrupted clip, $m=3$ of the $T=10$
tokens are sampled uniformly without replacement. Each selected token independently
receives one of
\begin{align}
\text{noise }(p=0.25):\quad&
x_t\leftarrow x_t+
\frac{0.5\lVert x_t\rVert_2}{\sqrt{d_{\mathrm{in}}}}\eta,
\quad \eta\sim\mathcal N(0,I),\\
\text{zero-fill }(p=0.50):\quad&
x_t\leftarrow\mathbf 0,\\
\text{local smoothing }(p=0.25):\quad&
x_t\leftarrow |\mathcal W_t|^{-1}\!\sum_{u\in\mathcal W_t}x_u,
\quad \mathcal W_t=\{u:|u-t|\le1\}\cap[1,T].
\end{align}
Selected positions receive the TRS target $y_t=1$; clean clips receive all-zero
targets.

\paragraph{Evaluation corruption.}
At high severity, Mask zeroes $\mathrm{round}(0.6T)=6$ sampled tokens; Noise
perturbs every token with relative standard deviation $\sigma=1.0$ under the
same norm-scaled Gaussian rule; Blur applies a width-$7$ symmetric moving
average with replicate padding; and Combined applies Noise, Blur, and Mask
sequentially. Training smoothing uses width $3$, so standalone Blur measures
severity transfer rather than a fully unseen mechanism.

For the implementation-level definitions below, token indices are zero-based.
Drop, Jitter, and Freeze use $n=\mathrm{round}(0.6T)=6$ at high severity.
Drop samples $n$ indices without replacement, sorts them, and processes them
sequentially:
\begin{equation}
\tilde{x}_t=
\begin{cases}
\tilde{x}_{t-1}, & t>0,\\
x_1, & t=0\ \text{and}\ T>1,\\
x_0, & T=1.
\end{cases}
\label{eq:drop_exact}
\end{equation}
Jitter performs $n$ independently sampled pairwise swaps with
$i,j\sim\mathrm{Uniform}\{0,\ldots,T-1\}$; repeated indices are permitted.
Freeze samples a start index
$a\sim\mathrm{Uniform}\{0,\ldots,T-n-1\}$ when $T-n>0$ (otherwise $a=0$)
and replaces positions $a,\ldots,\min(a+n-1,T-1)$ by $x_a$.
None of these three mechanisms appears during training.

\paragraph{Control implementations.}
FrameDrop~\cite{temprobust} uses the same backbone and optimiser with TRS
disabled. Each training clip is clean with probability $0.30$; otherwise,
$n_{\mathrm{drop}}=3$ tokens are sampled uniformly without replacement and
replaced by
\begin{equation}
\tilde{x}_t=x_{\max(0,t-1)}.
\label{eq:framedrop_exact}
\end{equation}
Thus a sampled token at $t=0$ is unchanged by the replacement rule.
For the frequency-only graph, let
$P(v)=\sum_n N_{v,n}/N$ and $P(n)=\sum_v N_{v,n}/N$. We remove pair-specific
association and use
\begin{equation}
G^{\mathrm{freq}}_{v,n}
=
\operatorname{clip}_{[-3,3]}
\left[
\log(P(v)+10^{-9})+\log(P(n)+10^{-9})
\right]
\label{eq:freq_exact}
\end{equation}
inside the same decoder with $\lambda=2$. The shuffled control row-permutes the
PMI graph, preserving its value distribution while destroying verb-specific
compatibility.

\paragraph{Calibration and repeatability.}
Because the selected $\lambda=2$ lies at the upper edge of the tested grid, we
treat it as a practical calibrated setting rather than evidence that the global
optimum is identified. All main-table methods share the train/validation
partition and a fixed evaluation-corruption seed. Multi-seed experiments use training seeds
$\{0,1,2\}$. Training corruption uses Python's \texttt{random} module for
index/operator sampling and PyTorch for Gaussian noise; bitwise determinism
across hardware was not enforced.

\subsection{Main Results}

Table~\ref{tab:main} reports all six EGTEA corruptions.
The clean baseline collapses under evidence-removing corruption, reaching 9.4\% on Mask and 6.8\% on Combined, showing that clean-input performance does not imply robustness.
TRS+CA+RVG achieves the highest AvgC (29.1\%) and RR (88.2\%), with its largest gains occurring on Mask and Combined.
These corruptions remove or compound temporal evidence, directly exposing both failure modes targeted by the framework: TRS limits the propagation of unreliable tokens, while RVG corrects unstable verb--noun combinations produced under ambiguous evidence.
On corruptions that preserve more usable temporal evidence, the margin narrows because the clean encoder can still recover a sufficiently informative representation.
The full method retains comparable clean accuracy (33.0\% versus 33.6\%), indicating that the robustness gains do not require a substantial clean-performance trade-off.

\begin{table}[t]
\centering
\caption{EGTEA top-1 action accuracy (\%), high severity. Drop, Jitter, and Freeze use mechanisms absent from training. AvgC averages the six corruptions; RR$=100\times$AvgC/Cl. FrameDrop is a temporal-augmentation baseline~\cite{temprobust}. Control rows were run on Mask, Noise, and Combined only. \textbf{Bold}: best per column among the full methods.}
\label{tab:main}
\small
\setlength{\tabcolsep}{3pt}
\resizebox{\linewidth}{!}{%
\begin{tabular}{lccccccc|cc}
\toprule
Method & Cl & Mask & Noise & Comb & Drop & Jitter & Freeze & AvgC & RR \\
\midrule
Clean baseline    & 33.6 & 9.4  & 23.4 & 6.8  & 24.4 & 27.0 & 32.0 & 20.5 & 61.0 \\
FrameDrop         & 32.7 & 9.6  & 23.4 & 7.6  & 23.7 & 26.1 & 30.5 & 20.1 & 61.6 \\
FrameDrop + RVG   & \textbf{36.5} & 12.7 & \textbf{30.6} & 11.3 & \textbf{28.5} & 29.4 & \textbf{35.2} & 24.6 & 67.5 \\
Corruption-Augmented (CA) & 28.4 & 22.1 & 19.7 & 20.7 & 22.0 & 25.5 & 25.5 & 22.6 & 79.6 \\
CA + RVG          & 31.3 & 25.2 & 25.1 & 25.5 & 24.4 & 28.1 & 29.7 & 26.3 & 84.0 \\
TRS + CA          & 29.5 & 23.6 & 22.9 & 24.0 & 24.3 & 26.9 & 27.5 & 24.9 & 84.4 \\
TRS + CA + RVG    & 33.0 & \textbf{27.8} & 29.6 & \textbf{28.0} & 27.5 & \textbf{31.1} & 30.4 & \textbf{29.1} & \textbf{88.2} \\
\midrule
TRS + CA + Shuffled  & 27.8 & 22.4 & 19.3 & 21.9 & -- & -- & -- & -- & -- \\
TRS + CA + Freq-only & 29.5 & 23.5 & 22.5 & 23.4 & -- & -- & -- & -- & -- \\
\bottomrule
\end{tabular}%
}
\end{table}

\paragraph{Component-wise contribution.}
The ablation pattern shows that robustness arises from complementary rather
than interchangeable gains.
Relative to CA (22.6\% AvgC), adding TRS raises performance to 24.9\%
($+2.3$ points), whereas adding RVG raises it to 26.3\% ($+3.7$ points).
Combining both reaches 29.1\%, improving by 4.2 points over TRS+CA and
2.8 points over CA+RVG.
These comparisons are descriptive rather than strictly additive because the
modules interact; nevertheless, they show that CA establishes the robustness
foundation, RVG provides the larger average correction, and TRS contributes
a smaller but complementary temporal improvement, with its largest gains on
Combined and Noise in this evaluation.

\paragraph{Controls.}
Averaged over Mask, Noise, and Combined, TRS+CA reaches 23.5\%; the true graph raises this to 28.5\%, whereas a row-shuffled graph lowers it to 21.2\%.
The shuffled graph preserves the weight distribution but destroys verb-specific compatibility, so its degradation indicates that the gain depends on where the pairwise structure is placed rather than on generic score perturbation.
Frequency-only yields 23.1\%, further ruling out marginal-frequency bias.
Table~\ref{tab:ablation} reproduces the ordering over three training seeds.

\begin{figure}[t]
\centering
\includegraphics[width=0.98\linewidth]{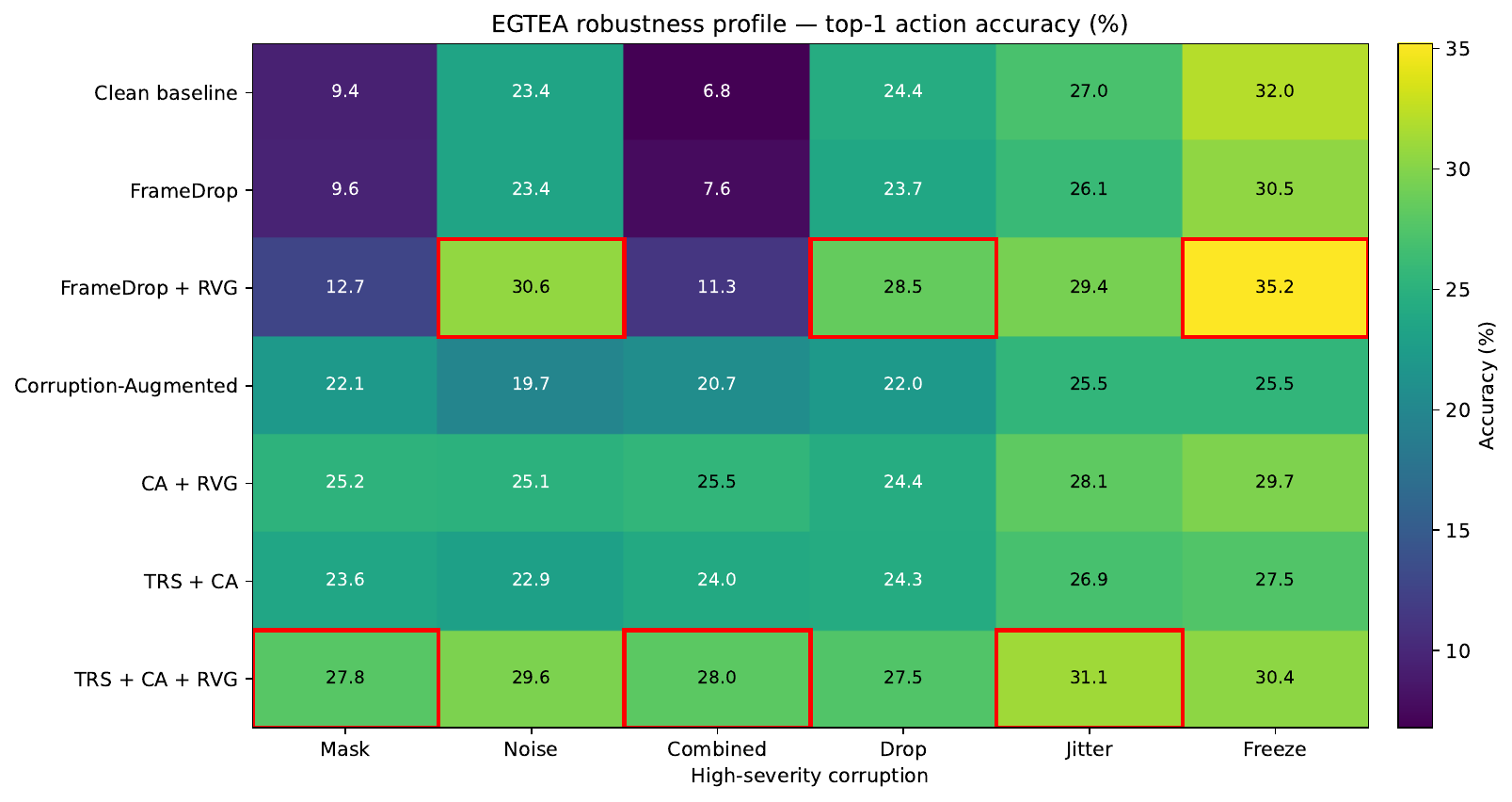}
\caption{EGTEA robustness profile (top-1 action accuracy \%; red: best per column). TRS+CA+RVG leads on Mask, Combined, and Jitter; FrameDrop+RVG leads on Noise, Drop, and Freeze.}
\label{fig:heatmap}
\end{figure}

\paragraph{No augmentation dominates every corruption.}
FrameDrop+RVG leads on Noise, Drop, and Freeze and has the highest clean accuracy, but collapses on Mask and Combined, giving a lower AvgC of 24.6\%.
This pattern is consistent with training-match effects: FrameDrop prepares the model for repetition-like degradation, while CA covers evidence removal and local smoothing.
Adding RVG lifts FrameDrop substantially but not to TRS+CA+RVG, showing that generic temporal augmentation does not remove the compositional failure mode.
Figure~\ref{fig:heatmap} visualises the full robustness profile.

\paragraph{Freeze: a mechanism-specific limitation of TRS.}
TRS+CA+RVG remains competitive on Freeze and improves over CA+RVG (30.4\% versus 29.7\%), but falls below the clean baseline.
Freeze creates an artefact-free yet temporally repetitive sequence: each frame looks plausible in isolation, while the corruption lies in the relationship between frames.
Because Eq.~\eqref{eq:head} reads frames independently, it has no explicit representation of repetition; a context-aware reliability estimator would be required to address this failure directly. A natural extension is therefore to replace the framewise suppression head with a lightweight context-aware estimator that compares adjacent tokens or models short temporal differences, enabling detection of repetition-based corruption without changing the main encoder.

\subsection{Transfer to EK100}

Table~\ref{tab:ek100} reports a within-pipeline comparison on EK100 rather than a leaderboard result.
The controls reproduce the EGTEA pattern: RVG improves both CA (6.7\% to 8.7\%) and TRS+CA (7.1\% to 8.2\%), while shuffled-PMI and frequency-only stay below the true graph.
RVG also improves the unseen-width Blur diagnostic on EGTEA by 4.1 points, indicating that the decision-level prior is not tied to the exact corruptions used for calibration.

Unlike EGTEA, adding TRS to CA+RVG lowers average corrupted accuracy from 8.7\% to 8.2\%.
A possible explanation is the substantially larger action space and greater semantic diversity of EK100, where correcting the many possible verb--noun combinations may be more valuable than identifying individual unreliable frames.
This remains a hypothesis because the datasets differ in more than label-space size.
We report the result without EK100-specific retuning, and interpret it as evidence that the relative importance of temporal and compositional robustness is dataset-dependent.

\begin{table}[t]
\centering
\caption{EK100 top-1 action accuracy (\%). Avg averages Mask, Noise, and Combined. $\lambda=2$ is transferred from EGTEA without re-tuning. \textbf{Bold}: best per column.}
\label{tab:ek100}
\small
\setlength{\tabcolsep}{6pt}
\begin{tabular}{lccccc}
\toprule
Method & Cl & Mask & Noise & Comb & Avg \\
\midrule
Clean baseline       & 8.9 & 0.5 & 5.8 & 0.5 & 2.3 \\
Corruption-Augmented (CA) & 8.2 & 6.9 & 6.2 & 6.9 & 6.7 \\
CA + RVG             & 9.4 & \textbf{9.0} & 8.3 & \textbf{8.7} & \textbf{8.7} \\
TRS + CA             & 9.4 & 8.4 & 6.3 & 6.7 & 7.1 \\
TRS + CA + RVG       & \textbf{10.2} & 8.1 & \textbf{8.5} & 8.0 & 8.2 \\
\midrule
TRS + CA + Shuffled  & 9.4 & 7.7 & 5.9 & 6.7 & 6.8 \\
TRS + CA + Freq-only & 9.4 & 7.0 & 5.5 & 7.3 & 6.6 \\
\bottomrule
\end{tabular}
\end{table}

\subsection{Multi-Seed Stability}

Table~\ref{tab:ablation} reports EGTEA Combined accuracy over three training seeds.
The full method has the highest mean with a small spread, and its margin over shuffled-PMI is much larger than seed-level variance.
The shuffled graph falls below the no-graph TRS+CA configuration on every seed, indicating that incorrect pairwise structure is actively harmful rather than merely unhelpful.

\begin{table}[t]
\centering
\caption{Multi-seed ablation on EGTEA Combined accuracy (mean$\pm$std, 3 seeds).}
\label{tab:ablation}
\small
\begin{tabular}{lc}
\toprule
Method & Combined accuracy \\
\midrule
CA                    & $0.201 \pm 0.003$ \\
CA + RVG              & $0.247 \pm 0.002$ \\
TRS + CA              & $0.238 \pm 0.008$ \\
TRS + CA + RVG        & $\mathbf{0.271 \pm 0.004}$ \\
TRS + CA + Shuffled   & $0.219 \pm 0.003$ \\
\bottomrule
\end{tabular}
\end{table}

\subsection{Diagnostic: TRS Response to Synthetic Masking}
\label{sec:diag}

Under synthetic masking, the head assigns mean suppression scores of 0.13 to clean frames and 0.81 to masked frames, giving frame-level AUC 1.00.
This verifies that the head uses the intended local corruption cue, but the perfect Mask AUC should not be interpreted as perfect corruption detection in general because zero-filled tokens provide a particularly distinctive signature.
Noise and relational corruptions such as Freeze are harder for a framewise estimator.
The signal transfers to EK100, where the corresponding means are 0.121 and 0.995, even though this does not produce the best corrupted accuracy there.
Detecting corruption and benefiting from that detection are therefore distinct questions.
Together, these diagnostics increase confidence that the observed improvements arise from the intended reliability-estimation mechanism rather than incidental regularisation.

\begin{figure}[t]
\centering
\includegraphics[width=0.86\linewidth]{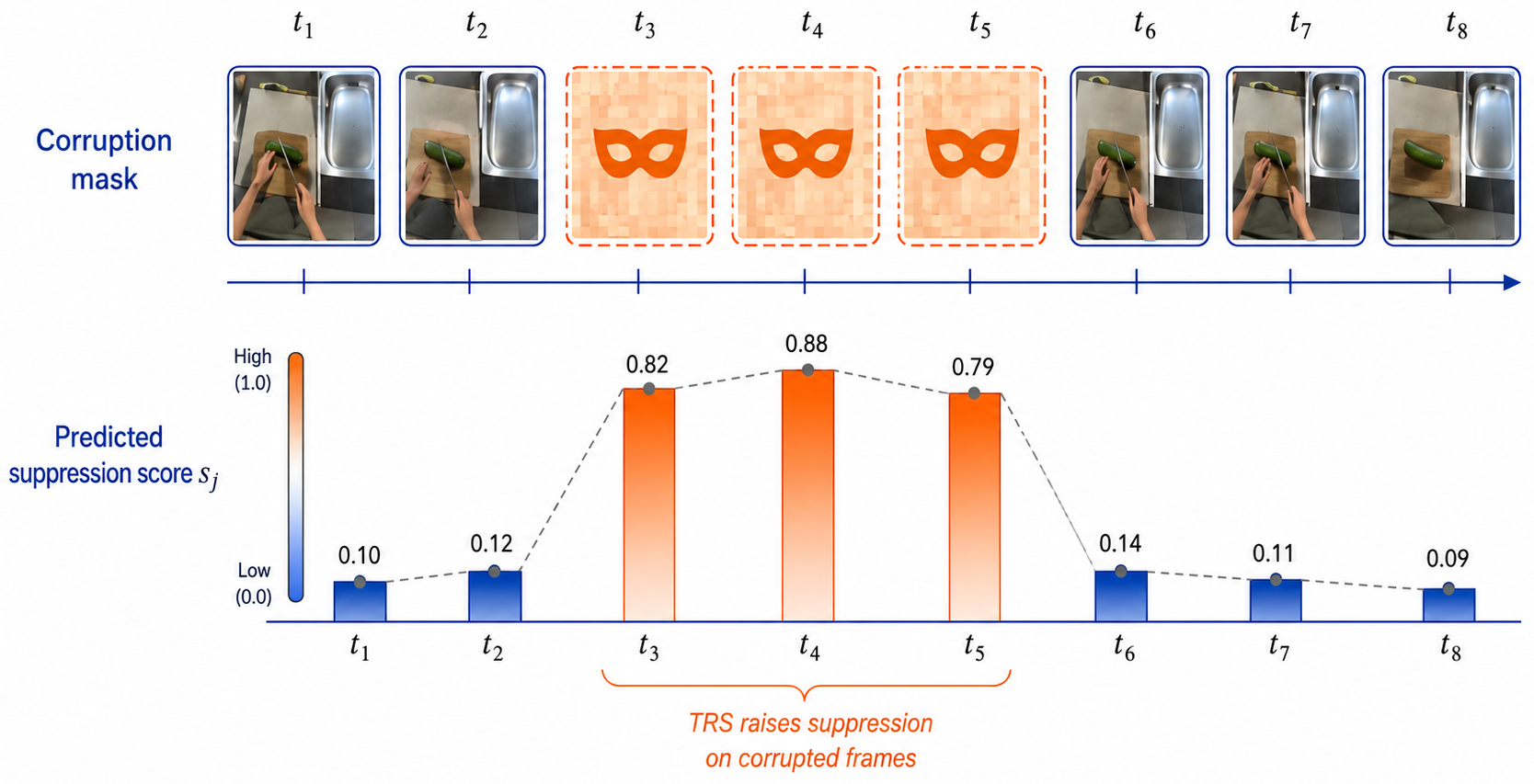}
\caption{TRS suppression scores on an EGTEA clip. Scores rise on masked frames $t_3$--$t_5$ and remain low elsewhere.}
\label{fig:trs_timeline}
\end{figure}

\subsection{TRS Control Ablation}

Random suppression reaches 22.8\% on Mask and 23.8\% on Combined, while TRS+CA reaches 23.6\% and 24.0\%; ground-truth-mask suppression reaches 24.2\% and 25.1\%.
Random suppression remains competitive on Combined because broadly removing evidence can reduce contamination when corruption is widespread, whereas learned suppression is more useful on localised Mask corruption.
With RVG, TRS+CA+RVG reaches 27.8\%/28.0\% and ground-truth-mask suppression reaches 27.4\%/28.2\%.
A hard binary mask is not an upper bound for a model trained with continuous scores, so we avoid calling it an oracle.
Overall, TRS provides a modest complementary temporal gain, while RVG accounts for the larger semantic correction.

\subsection{What RVG Actually Changes}

\begin{figure}[t]
\centering
\includegraphics[width=0.86\linewidth]{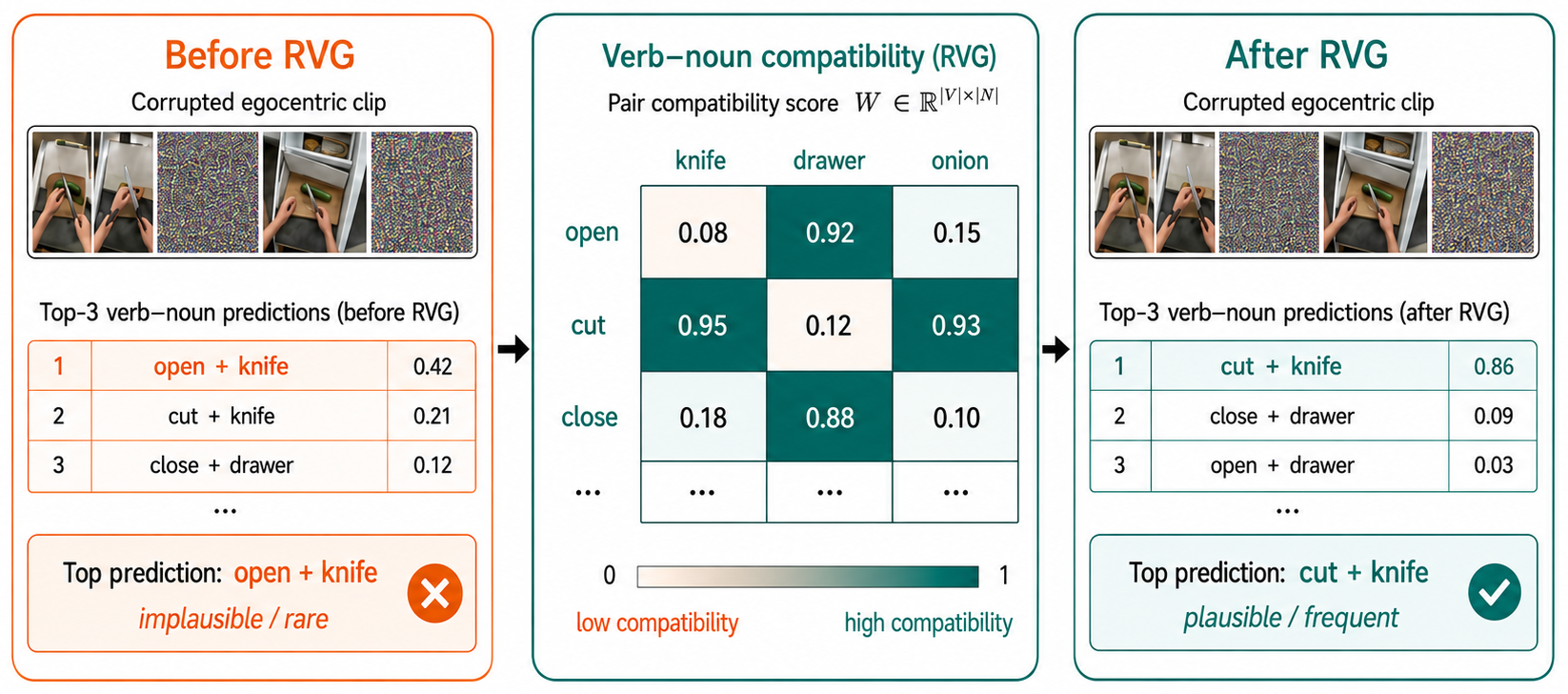}
\caption{Qualitative RVG correction: graph re-scoring replaces the low-support \textit{open knife} prediction with \textit{cut knife}.}
\label{fig:rvg_qual}
\end{figure}

Under Combined corruption, RVG raises the mean PMI of predicted pairs from 0.90 to 1.42 and reduces rare predictions from 15.4\% to 1.1\%.
Because these quantities are derived from the same co-occurrence graph, we treat them as consistency checks; the accuracy gains remain the primary evidence. We use the count threshold $5$ for graph abstention and the rare-prediction diagnostic, but $30$ for the rare-ground-truth analysis because only 78 validation examples fall below the latter threshold and too few fall below five for a stable conditional-accuracy comparison.
Figure~\ref{fig:rvg_qual} shows a representative correction.

Table~\ref{tab:verb_noun} separates verb, noun, and joint accuracy on a separately logged full-validation prediction file.
RVG changes decoded verb and noun correctness only modestly, yet raises joint action accuracy by 4.9 points.
This is the expected signature of compositional correction: the individual components change little while their pairing improves.
RVG also improves both seen and rare ground-truth pairs, indicating that suppressing rare predictions does not come at the cost of rare-but-correct targets.

\begin{table}[t]
\centering
\caption{Left: verb, noun, and joint accuracy under EGTEA Combined corruption. Right: accuracy on seen versus rare ground-truth pairs (rare = training count $<30$, $n=78$ of 2022).}
\label{tab:verb_noun}
\small
\setlength{\tabcolsep}{5pt}
\begin{tabular}{lccc|ccc}
\toprule
& \multicolumn{3}{c|}{Per-stream} & \multicolumn{3}{c}{By GT rarity} \\
Method & Verb & Noun & Action & All & Seen-GT & Rare-GT \\
\midrule
TRS + CA       & 36.6 & 37.8 & 22.5 & 22.2 & 22.8 & \phantom{0}6.4 \\
TRS + CA + RVG & 37.9 & 37.1 & \textbf{27.4} & \textbf{28.3} & \textbf{28.9} & \textbf{15.4} \\
\bottomrule
\end{tabular}
\end{table}

\subsection{Limitations}
\label{sec:limitations}

A central limitation is that TRS relies on exact synthetic frame-level corruption labels during training.
Although inference is label-free, obtaining equivalent supervision for naturally occurring degradation remains open; practical systems may require self-supervised anomaly detection, weak supervision, or proxy reliability signals.
The framework also operates on pre-extracted 2016-era TSN features and applies corruption at the feature level rather than to raw sensor video.
Consequently, the experiments do not establish robustness to real motion blur, rolling-shutter artefacts, packet loss, or other sensor-specific degradation, nor do they demonstrate on-device performance.
The suppression head estimates each frame independently and cannot directly model degradation expressed only through inter-frame relationships; Freeze is the clearest example. Moreover, the TRS controls evaluate attention suppression and reliability-weighted pooling jointly, so their individual contributions remain unresolved without additional training runs.
RVG depends on training-set co-occurrence and abstains where pair support is too low, so it may contribute less under severe label-distribution shift or personalised action vocabularies.
The graph weight is chosen in a preliminary sweep on the EGTEA validation split rather than through a fully nested calibration protocol; because $\lambda=2$ is also the largest tested value, future work should confirm it using an independent calibration partition and a broader grid. Finally, evaluation is limited to one backbone, feature-level corruptions, and a restricted EK100 transfer protocol; raw-video evaluation, additional architectures, and in-the-wild wearable streams remain important future work.

\section{Conclusion}

We studied robust egocentric action anticipation through two complementary failure modes: unreliable temporal evidence during encoding and low-support verb--noun composition during decoding.
TRS reduces the influence of corrupted observations during attention and pooling, while RVG uses training-set compatibility structure to correct unstable action compositions.
Their combination achieves the strongest average corrupted performance and relative robustness on EGTEA, while the EK100 results show that the relative importance of temporal reliability and compositional correction is dataset-dependent.
More broadly, our results suggest that robust wearable action anticipation is not solely a representation-learning problem: it requires both reliable temporal evidence and structured semantic reasoning.
We hope that the proposed corruption protocol and diagnostic analyses encourage future work on deployment-oriented robustness for proactive egocentric AI.

\bibliographystyle{splncs04}
\bibliography{refs}

\end{document}